\pdfoutput=1
\documentclass{article}
\usepackage{graphicx} 

\usepackage[preprint]{neurips_2023}

\usepackage[utf8]{inputenc} % allow utf-8 input
\usepackage[T1]{fontenc}    % use 8-bit T1 fonts
\usepackage{hyperref}       % hyperlinks
\usepackage{url}            % simple URL typesetting
\usepackage{booktabs}       % professional-quality tables
\usepackage{amsfonts}       % blackboard math symbols
\usepackage{nicefrac}       % compact symbols for 1/2, etc.
\usepackage{microtype}      % microtypography
\usepackage{xcolor}         % colors
\usepackage{subfigure}

\title{Drive Like a Human: Rethinking Autonomous Driving with Large Language Models}

\author{%
Daocheng Fu$^{1}$\footnotemark[1] \quad  Xin Li$^{1,2}$\footnotemark[1] \quad  Licheng Wen$^{1}$\footnotemark[1] \quad  Min Dou$^{1}$\quad Pinlong Cai$^{1}$\\ \textbf{Botian Shi}$^{1}$\footnotemark[2]\quad \textbf{Yu Qiao}$^{1}$\\
$^{1}$Shanghai AI Lab, $^{2}$East China Normal University \\
\{fudaocheng, lixin, wenlicheng, doumin, caipinlong, shibotian, qiaoyu\}@pjlab.org.cn
}

\begin{document}

\maketitle
\renewcommand{\thefootnote}{\fnsymbol{footnote}} 
\footnotetext[1] {Equal contribution. Sorted alphabetically by surname.}
\footnotetext[2] {Corresponding author.}

\begin{abstract}

In this paper, we explore the potential of using a large language model (LLM) to understand the driving environment in a human-like manner and analyze its ability to reason, interpret, and memorize when facing complex scenarios. We argue that traditional optimization-based and modular autonomous driving (AD) systems face inherent performance limitations when dealing with long-tail corner cases. To address this problem, we propose that an ideal AD system should drive like a human, accumulating experience through continuous driving and using common sense to solve problems. To achieve this goal, we identify three key abilities necessary for an AD system: reasoning, interpretation, and memorization. We demonstrate the feasibility of employing an LLM in driving scenarios by building a closed-loop system to showcase its comprehension and environment-interaction abilities. Our extensive experiments show that the LLM exhibits the impressive ability to reason and solve long-tailed cases, providing valuable insights for the development of human-like autonomous driving.
The related code are available at \url{https://github.com/PJLab-ADG/DriveLikeAHuman}.

\end{abstract}

\section{Introduction}
\label{sec:intro}

Imagine if you are sitting behind the wheel waiting for the green light in front of a stop sign. Meanwhile, a pickup truck carrying traffic cones is crossing the intersection ahead. As a human driver, you can leverage your common sense knowledge to reason that \emph{these traffic cones are cargo on a pickup truck, and it doesn't mean the road is under construction}. However, these scenarios, which are easy for human drivers to handle, are long-tail corner cases for many existing autonomous driving (AD) systems~\cite{chen2022milestonessurvey,chen2023milestonespart1,chen2023milestonespart2}. 
Although self-driving developers can address this case by hand-crafting rules or collecting more data specifically about \emph{traffic cones on vehicles instead of on the ground} to prevent sudden braking, the algorithm would fail in the opposite cases when encountering on the ground that marks a no-go zone. It's like solving one problem only to find another cropping up, especially with the infinite rare cases in the real world that are even beyond our imagination. This is why we believe that traditional optimization-based and modular AD systems inherently face performance bottlenecks~\cite{chen2022milestonessurvey,uniad}.

\begin{figure}[!tbp]
\centering
\includegraphics[width=0.98\linewidth]{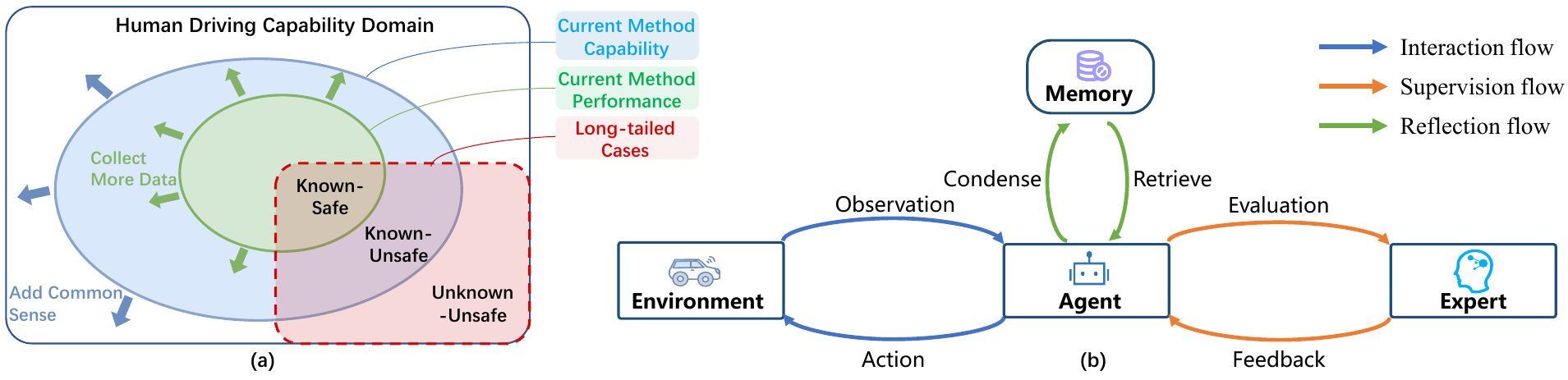}
 \caption{(a) The relationships between human driving and existing autonomous driving systems, especially highlights the limitations of current approaches and why they cannot solve all long-tail cases. (b) The schema of a system that can drive like a human. The agent can explore and interact with the environment and conduct self-reflection based on the expert's feedback, finally accumulating the experience.}
 \label{fig:motivation}
\end{figure}

We rethink the storyboard of autonomous driving and clarify why conventional optimization-based AD systems struggling with the challenging open world in Fig. \ref{fig:motivation} (a). Although the system that built upon the optimization theory can easily divide the complex autonomous driving task into a set of sub-tasks. The goal of optimizing the loss function tends to be trapped by local optimizations when facing a complex scenario, which limits its generalization ability. Incorporating more data (green arrows in graph) can only reduce the performance gap between to current model (green ellipse) and the maximum capacity of optimization-based methods (blue ellipse). This is mainly because the optimization process focuses on learning the dominant patterns within the data, often overlooking infrequent long-tail corner cases. Without the incorporation of common sense (blue arrows), the capacity of the model (blue ellipse) cannot be prompted.

Moreover, there are always inexhaustible unknown long-tail cases during the continuous data collection. Compared with current solutions struggling to cope with these long-tail corner cases, humans always can solve them with skill and ease via their experiences and common sense. A straightforward idea comes out: \textbf{Is it possible for us to make such a system that can drive like a human who can accumulate experience through continuous driving instead of fitting the limited training corpus?}

According to the recent research~\cite{surveyembodiedai,ramakrishnan2021exploration,zhu2023excalibur,wang2023voyager,zhu2023ghost}, we think the previous modular AD systems can be regarded as an Internet AI~\cite{chen2023milestonespart1,chen2023milestonespart2,li2023logonet,uniad} that is trained on task-specific corpus without advanced intelligence like reasoning, interpretation and self-reflection. We claim that it is necessary to borrow ideas from Embodied AI~\cite{pfeifer2004embodied,smith2005development} research if we want to get an agent that can drive a car like an experienced human driver. Human learns to drive from interaction with the real environment and gets feedback to refine the road sense by explaining, reasoning, and condensing the memory of various scenarios and the corresponding operations. 
Furthermore, owing to their logical reasoning ability, human drivers can use their common sense to summarize rules and apply them in more general scenarios (inductive reasoning)\cite{hayes2010inductive}. Meanwhile, the previous experience can be aroused subconsciously to handle unpredictable scenarios (deductive reasoning)\cite{johnson1999deductive}.

Towards the goal of driving like a human, we identify three abilities that are necessary: 1) \textbf{Reasoning}: Given a specific driving scenario, the model should be able to make decisions by reasoning via common sense and experience. 2) \textbf{Interpretation} The decisions made by the agent should able to be interpreted. This demonstrates the ability of introspection and the existence of declarative memory. 3) \textbf{Memorization}: After reasoning and interpreting scenarios, a memory mechanism is required to remember previous experiences and enable the agent to make similar decisions once facing similar situations.

Based on the above three properties, we refer to the paradigm of human learning to drive and condense a canonical form of a driving system depicted in Fig. \ref{fig:motivation} (b). This schema includes four modules: (1) Environment creates a stage that the agent can interact with by the interaction flow; (2) Agent stands for a driver that can perceive the environment and make decisions utilizing its memory and learning from expert advice; (3) Memory allows the agent to accumulate experience and perform actions with it via the reflection flow; and (4) Expert provides advice on agent training and gives feedback when it acts inconsistently, which forms the supervision flow.
To be specific, as a universal driving framework, the Environment, the Agent, and the Expert can be represented by the real world or simulator, human drivers or driving algorithms, and simulator or instructor feedback, respectively.

Inspired by recent research, the large language model (LLM) can be considered as an early version of Artificial General Intelligence (AGI)\cite{bubeck2023sparks,wang2023voyager,zhu2023ghost,ahn2022can,shinn2023reflexion}, owing to its remarkable emergent ability~\cite{wei2022emergent} and new techniques like Instruct Following\cite{ouyang2022training} and In-Context Learning (ICL)\cite{brown2020language}. The extensive experimental results of recently released LLMs like ChatGPT~\cite{chatgpt} have demonstrated their ability of reasoning, interpretation, and memorization \cite{weng2023prompt}.
Therefore, in this paper, we try to initially explore the ability of LLM to understand driving traffic scenes like humans and analyze the LLM's ability of reasoning, interpretation and memorization in handling scenarios like long-tail corner cases through a series of qualitative experiments. 
Specifically, we first build a closed-loop system to demonstrate the comprehension and environment-interaction ability of an LLM (GPT-3.5) in driving scenarios. Then we demonstrate the reasoning and memorization abilities by solving several typical long-tailed cases that are difficult for modular AD systems to handle while easy for human drivers.

The main contributions of this paper are as follows:
\begin{enumerate}
    \item We dive deep into how to make autonomous driving systems drive like a human to prevent catastrophic forgetting of the existing AD systems when facing long-tail corner cases and summarize into three key abilities to drive like a human: Reasoning, Interpretation and Memorization.
    \item We are the first to demonstrate the feasibility of employing LLM in driving scenarios and exploit its decision-making ability in the simulated driving environment.
    \item Extensive experiments in our study express impressive comprehension and the ability to solve long-tailed cases. We hope that these insights will inspire academia and industry to contribute to the development of human-like autonomous driving.
\end{enumerate}

\section{Closed-loop interaction ability in driving scenarios}

\begin{figure}[tbp]
    \centering
    \includegraphics[width=\textwidth]{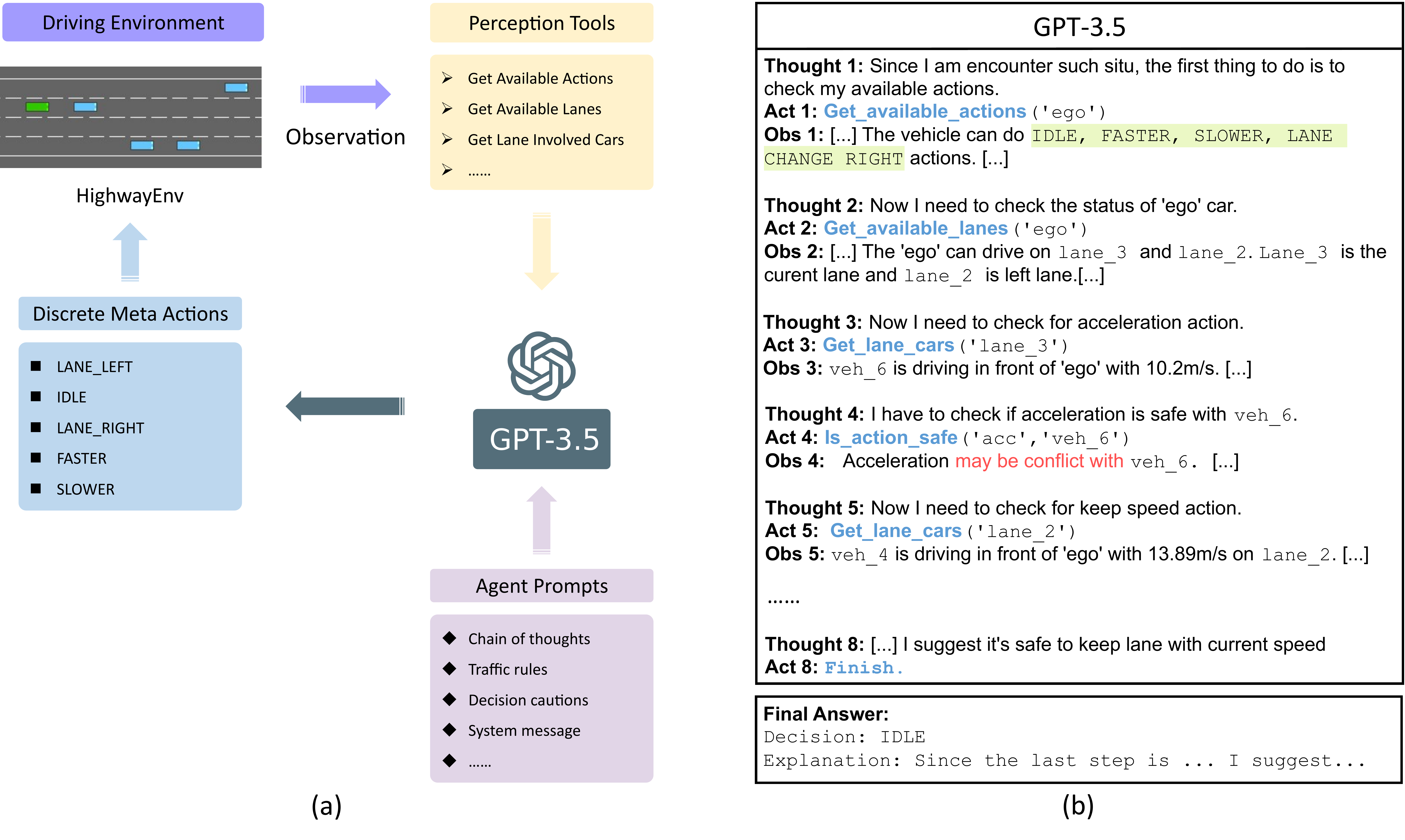}
    \caption{GPT-3.5 closed-loop driving in HighwayEnv: (a) GPT-3.5 observes its environment in HighwayEnv using perception tools and makes decisions to control vehicles, forming a closed loop. (b) GPT-3.5 employs the ReAct strategy to plan actions and use tools, while perceiving its surroundings through a cycle of thought, action, and observation}
    \label{fig:closeLoop}
\end{figure}

The ability to interpret allows LLM to understand its driving environment, forming the basis for its interaction with environments, and empowering its reasoning, and memorization abilities. 
We conducted a closed-loop driving experiment on HighwayEnv\footnote{\href{https://github.com/Farama-Foundation/HighwayEnv}{https://github.com/Farama-Foundation/HighwayEnv}} using GPT-3.5 to verify LLM’s interpretation and environmental interaction abilities. As a text-only large language model, GPT-3.5 cannot directly interact with HighwayEnv, so we provided perception tools and agent prompts to aid its observation and decision-making. As shown in Fig. \ref{fig:closeLoop}, the Agent Prompts provide GPT-3.5 with information about its current actions, driving rules, and cautions.
GPT-3.5 employs the ReAct strategy \cite{yao2022react} to perceive and analyze its surrounding environment through a cycle of thought, action, and observation.
% while Perception Tools parse observations generated by HighwayEnv to help GPT-3.5 understand the current scenario. 
Based on this information, GPT-3.5 makes decisions and controls vehicles in HighwayEnv, forming a closed-loop driving system.

Like humans, GPT-3.5 evaluates the potential consequences of its actions while driving and weighs the outcomes to make the most sensible decision. Unlike widely used Reinforcement Learning (RL)-based and Search-based approaches, GPT-3.5 not only interprets scenarios and actions but also utilizes common sense to optimize its decision-making process.

\begin{figure}[tbp]
    \centering
    \includegraphics[width=\textwidth]{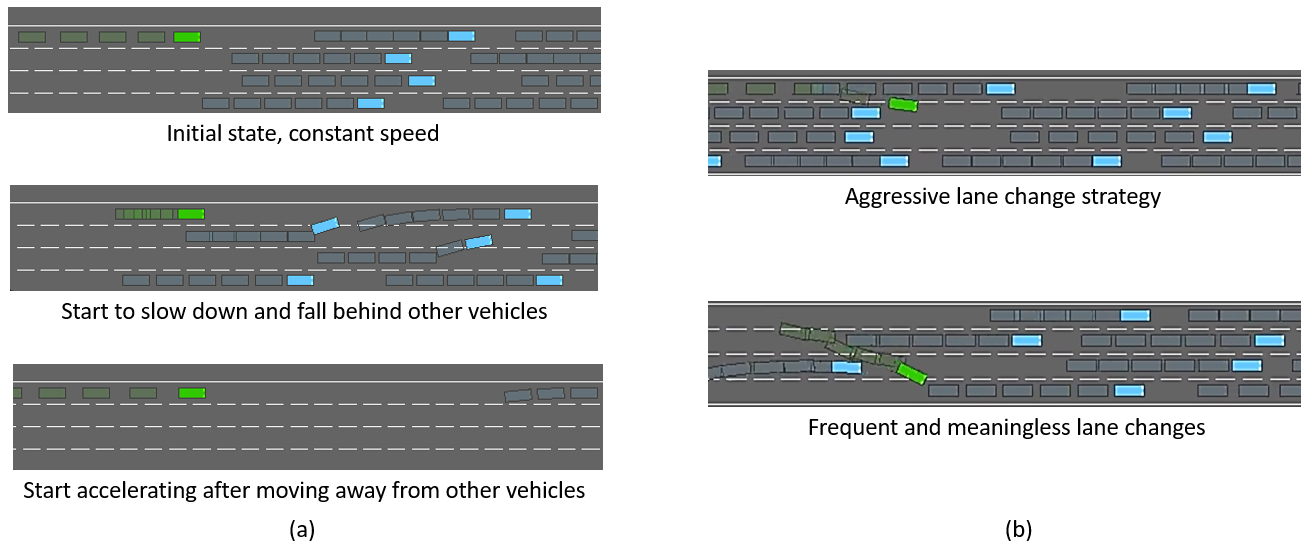}
    \caption{Driving behavior of RL-based and Search-based methods: (a) RL-based agents focus solely on achieving the final reward, disregarding the intermediate steps. This allows them to take unconventional actions, such as slowing down to fall behind other vehicles and then driving on an open road to avoid collisions. (b) Search-based methods make decisions by optimizing objective functions. They may pursue aggressive behavior by seeking max efficiency while ensuring safety.}
    \label{fig:RL&SearchBased}
\end{figure}

Compared to the RL-based approach, GPT-3.5 achieve a zero-shot pass rate of over 60\% in HighwayEnv without any fine-tuning. In contrast, the RL-based approach is heavily dependent on numerous iterations to achieve a competitive performance.
For example, as Fig.\ref{fig:RL&SearchBased} (a) shows, due to the severe penalty for collisions, the RL-based agent learned a policy that in order to prevent collisions it will slow down at the beginning to create an extensive space for the subsequent accelerating.
It shows that the RL-based approach often produces such unexpected solutions.

The search-based approach makes decisions by optimizing an objective function, ignoring undefined parts not mentioned in the function. As Fig.\ref{fig:RL&SearchBased} (b) shows, search-based agents may exhibit aggressive lane changes to achieve high driving efficiency, thereby increasing the risk of collisions. Additionally, the search-based approach may make meaningless lane changes even when no other vehicles are ahead. This may be because, for search-based agents, lane changing and maintaining speed have equal priority in the objective function under the premise of safety. As a result, the agent will randomly choose one of the actions. 

In summary, neither RL-based nor Search-based approaches can truly think and drive like humans because they lack common sense, the ability to interpret scenarios, and the ability to weigh the pros and cons.
In contrast, GPT-3.5 can explain the consequences of each action, and by providing prompts, we can make GPT-3.5 value-oriented and enable it to make more human-like decisions.

\begin{figure}[tbp]
    \centering
    \includegraphics[width=0.98\textwidth]{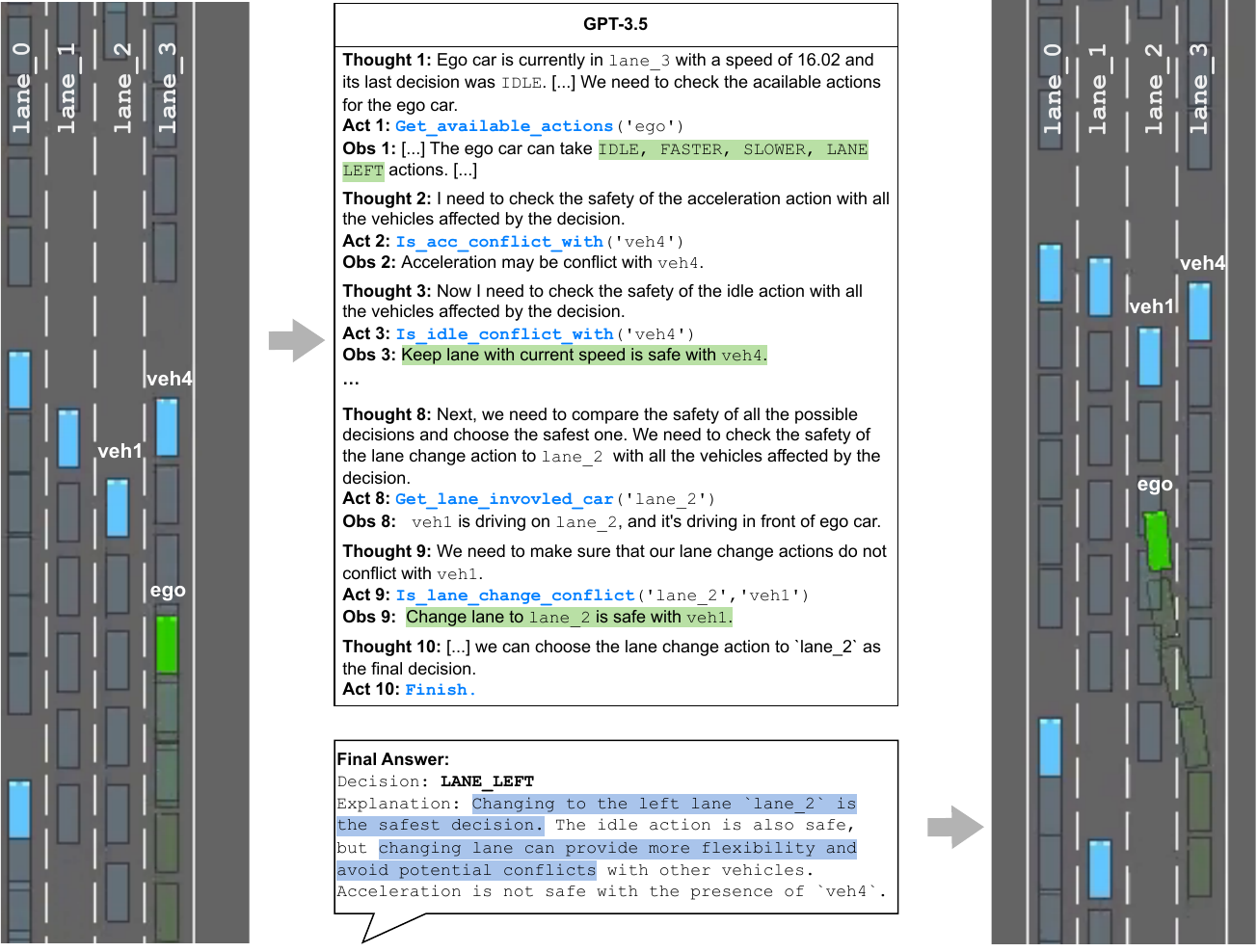}
    \caption{The lane-change decision-making process by GPT-3.5}
    \label{fig:close_case1}
    \vspace{-10pt}
\end{figure}

\begin{figure}[!b]
    \centering
    \includegraphics[width=0.98\textwidth]{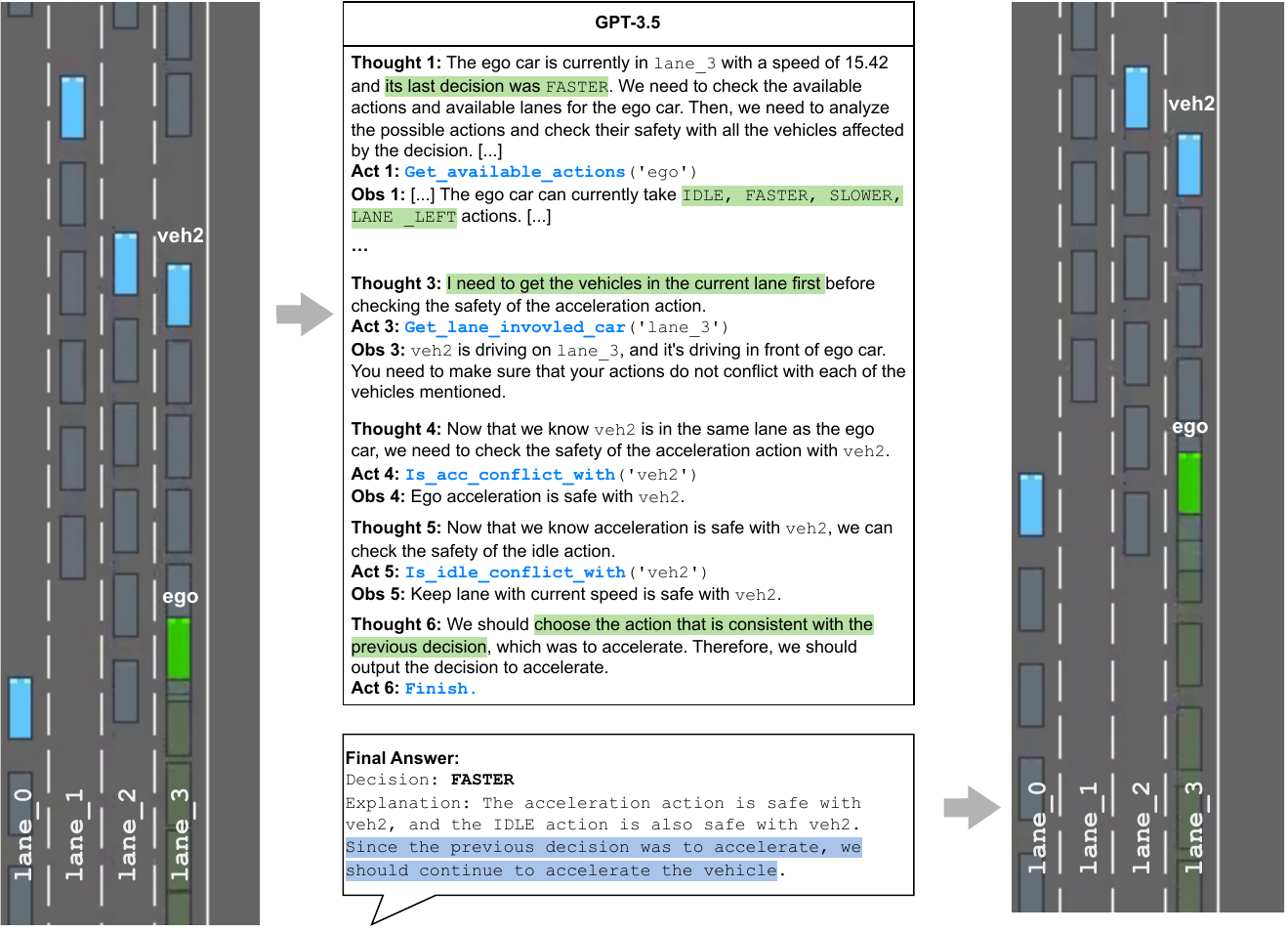}
    \caption{The decision consistency during acceleration process}
    \label{fig:close_case2}
\end{figure}

We present two examples that demonstrate the interpretation and interaction abilities of GPT-3.5 in the HighwayEnv environment, as well as its consistency in decision-making during the closed-loop process.
The first case is depicted in Fig.\ref{fig:close_case1}, involving a green ego car driving in the rightmost lane \texttt{lane\_3}. The ego car has been following its leading car \texttt{veh4} on \texttt{lane\_3} for a while, while \texttt{veh1} on the left lane \texttt{lane\_2} has been travelling at a faster speed than \texttt{veh4}.
GPT-3.5 then begins its ReAct process. It first determines the actions that the ego car can currently perform, including accelerating, decelerating, and maintaining speed in the current lane, as well as changing lanes to the left. However, since the ego car is in the rightmost lane, it cannot continue to change lanes to the right.
Next, GPT-3.5 checks the safety of each available action. The perception tools show that acceleration may cause a potential collision with the front car \texttt{veh4}, while maintaining speed is quite safe. When checking the change lane to the left action, GPT-3.5 first determines which vehicle will be affected by such action on \texttt{lane\_2}, and then learns that changing lanes to the left is safe with \texttt{veh1}.
At this point, GPT-3.5 has examined each action and made a final decision to change to the left lane. It provides a mature and reasonable explanation, stating that although the idle and change lane actions are both safe, changing to \texttt{lane\_2} is a better move since it provides more flexibility for the ego car. Considering that \texttt{veh1} has a faster speed, this decision can lead to better performance.

Closed-loop driving in HighwayEnv not only requires LLMs to make safe decisions at each time step but also demands consistency between decisions and avoiding behaviors such as frequent acceleration and deceleration and meaningless lane changes. 
In our framework, the decision result and explanation from the previous frame are included as part of the agent prompts and input into GPT-3.5. We use the second example in Fig. \ref{fig:close_case2} to demonstrate that GPT-3.5 has such decision-making consistency.

In this example, the green ego car is on the rightmost lane and is following \texttt{veh2} while maintaining a relatively long distance. In the previous decision, GPT-3.5 determined that the distance from the leading car was too far, so it decided to speed up to keep up with \texttt{veh2}.
At the beginning of the ReAct process, GPT-3.5 still uses the \texttt{Get\_available\_action} tool to obtain all four available actions in the current time step. It then finds out that \texttt{veh2} is still driving in front of the ego car, and both the idle and acceleration actions are safe with the leading vehicle.
The final decision made by GPT-3.5 is to keep accelerating, since it "chooses the action that is consistent with the previous decision," as explained in its final answer. As a result, the ego car shortens the distance with the front vehicle, which is more conducive to the overall traffic flow.
Compared with the first example, due to the reference to the previous decision result, the number of tools invoked by GPT-3.5 and the reasoning cost are significantly reduced.

\section{Reasoning ability with common sense}

While both human drivers and previous optimization-based AD systems possess basic driving skills, a fundamental difference between them is that humans have a common sense understanding of the world. 
Common sense is sound and practical judgement of what is going on around us, accumulated from daily lives \cite{wiki:commonsense}. The common sense that contributes to driving can be derived from all aspects of everyday life. 
When presented with a new driving situation, a human driver can quickly assess the scenario based on common sense and make a reasonable decision. 
In contrast, conventional AD systems may have experience in the driving domain, but they lack common sense and thus unable to tackle such situations. 
% And this issue persists no matter how much driving cases is added.

LLMs like GPT-3.5, have been trained on vast amounts of natural language data and are knowledgeable about common sense \cite{bian2023chatgpt}.
This marks a significant departure from traditional AD methods and empowers LLMs to reason through complex driving scenarios using common sense, much like human drivers.
In this section, we evaluate two typical long-tail cases in autonomous driving systems, which involves a pickup truck carrying traffic cones as described at the beginning of Section\ref{sec:intro}.

\begin{figure}[tbp]
	\centering
	\subfigure[Decision when the cones are loaded on the truck-bed.]{
		\begin{minipage}{\textwidth}
                        \includegraphics[width=\textwidth]{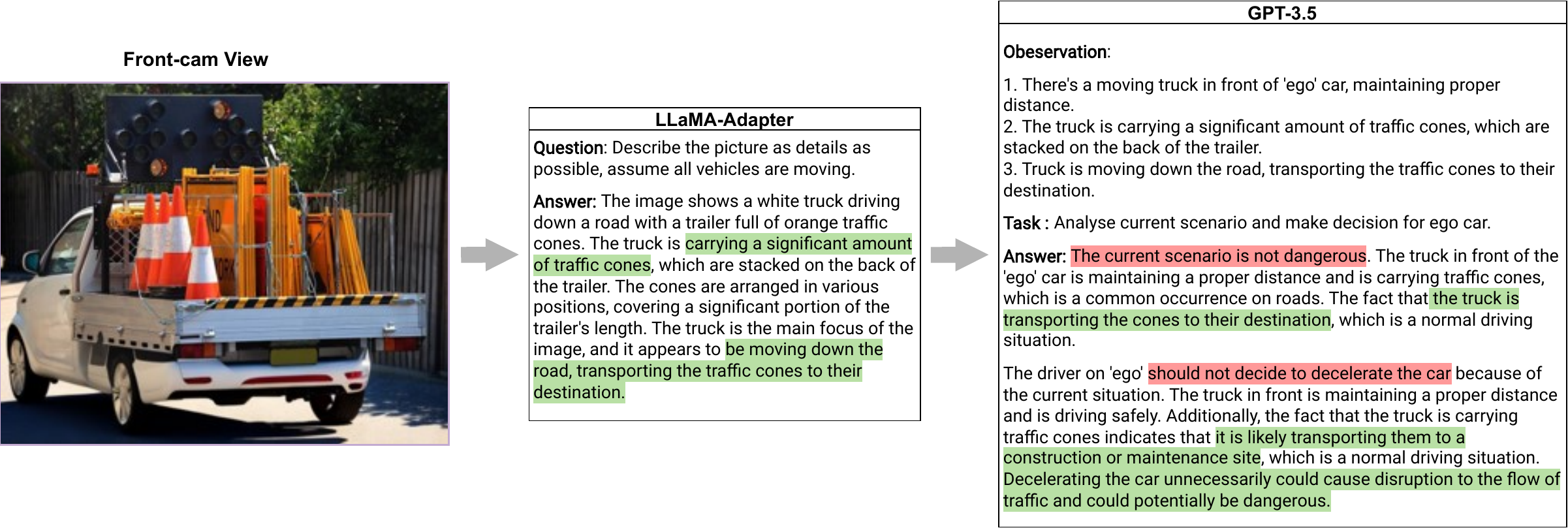} \\
		\end{minipage}
            \label{fig:cones_on_trailer}
	}
	\subfigure[Decision when the cones are scattered on the ground.]{
		\begin{minipage}{\textwidth}
			\includegraphics[width=\textwidth]{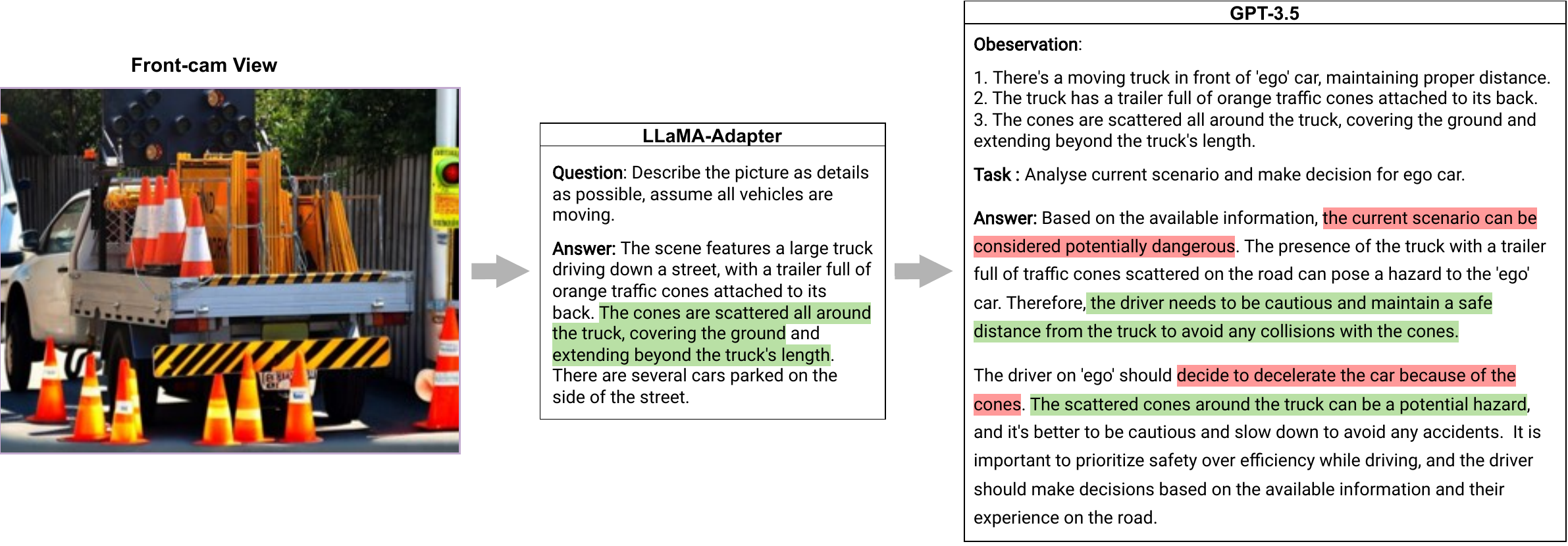} \\
		\end{minipage}
            \label{fig:cones_on_ground}
	}
	\caption{Two similar long-tail cases with a pickup truck carrying traffic cones.} 
	\label{fig:truck_with_cones}
\end{figure}

As depicted in Fig.\ref{fig:truck_with_cones}, two similar yet distinct photographs are fed into the LLMs. The first photo portrays a pickup truck carrying several traffic cones in its truck-bed en route to their destination. The second one also depicts a pickup with cones in its truck bed, yet with additional cones scattered on the ground in the surrounding area.
As GPT-3.5 lacks the ability to process multimodal inputs including images, we employ the LLaMA-Adapter v2 \cite{gao2023llamaadapter} as an image processing front-end. We instruct LLaMA-Adapter to describe the photo as details as possible.  This description is then utilized as an observation, and we ask GPT-3.5 to assess whether the scenario is potentially hazardous and to make a decision for the ego car, which is assumed to be following behind the truck.

In the first case depicted in Fig.\ref{fig:cones_on_trailer}, LLaMA-Adapter identified that the pickup truck in the photograph is carrying multiple traffic cones and inferred that it may be transporting them to their destination. Based on these observations, GPT-3.5 successfully analyzed the driving scenario. Instead of being misled by the presence of traffic cones, GPT-3.5 deemed this scenario to be non-hazardous, based on the common sense that a truck transporting its cargoes to their destination is a common occurrence.  GPT-3.5 advised the driver of the ego car that there was no need to slow down and cautioned that unnecessary slowdowns could potentially be dangerous for traffic flow.

For the second case,  depicted in Fig.\ref{fig:cones_on_ground}, traffic cones are not only inside the truck bed, but also scattered on the ground, which accurately represented by LLaMA-Adapter. Despite the minor difference from the first case, GPT-3.5's response is diametrically opposed. It considered this scenario to be potentially dangerous due to the scattered cones around the truck and advised the driver of the ego car to decelerate and maitain a distance to avoid any collisions with those cones.

The above examples demonstrate LLM's powerful zero-shot understanding and reasoning abilities in driving scenarioes. The utilization of common-sense knowledge not only allows LLMs to better comprehend the semantic information within the scenario but also enables them to make more rational decisions, which are more compatible with human driving behavior.
Therefore, possessing common-sense knowledge increases the upper limit of an autonomous driving system's capabilities, enabling it to handle unknown long-tail cases and truly approach the driving capabilities of human drivers.

\section{Performance enhancement through memorization ability}

Continuous learning~\cite{parisi2019continual} is another key aspect for human to drive. Novice drivers typically drive cautiously due to their limited experience when encountering complex traffic situations. Over time, as they gain more experience behind the wheel, drivers encounter new traffic scenarios, develop new driving skills, and consolidate their previous experience, eventually becoming experienced drivers.
The optimization-based approach aims to imitate the process of continuous learning by acquiring more and more failed cases and using them to retrain the neural network. However, this approach is not only tedious and expensive but also fundamentally incapable of achieving continuous learning. The distribution difference between regular driving and long-tail corner cases poses a significant challenge for balancing the two, eventually resulting in catastrophic forgetting. Therefore, a more effective approach is needed to achieve true continuous learning in autonomous driving systems.

\begin{figure}[tbp]
    \centering
    \includegraphics[width=\textwidth]{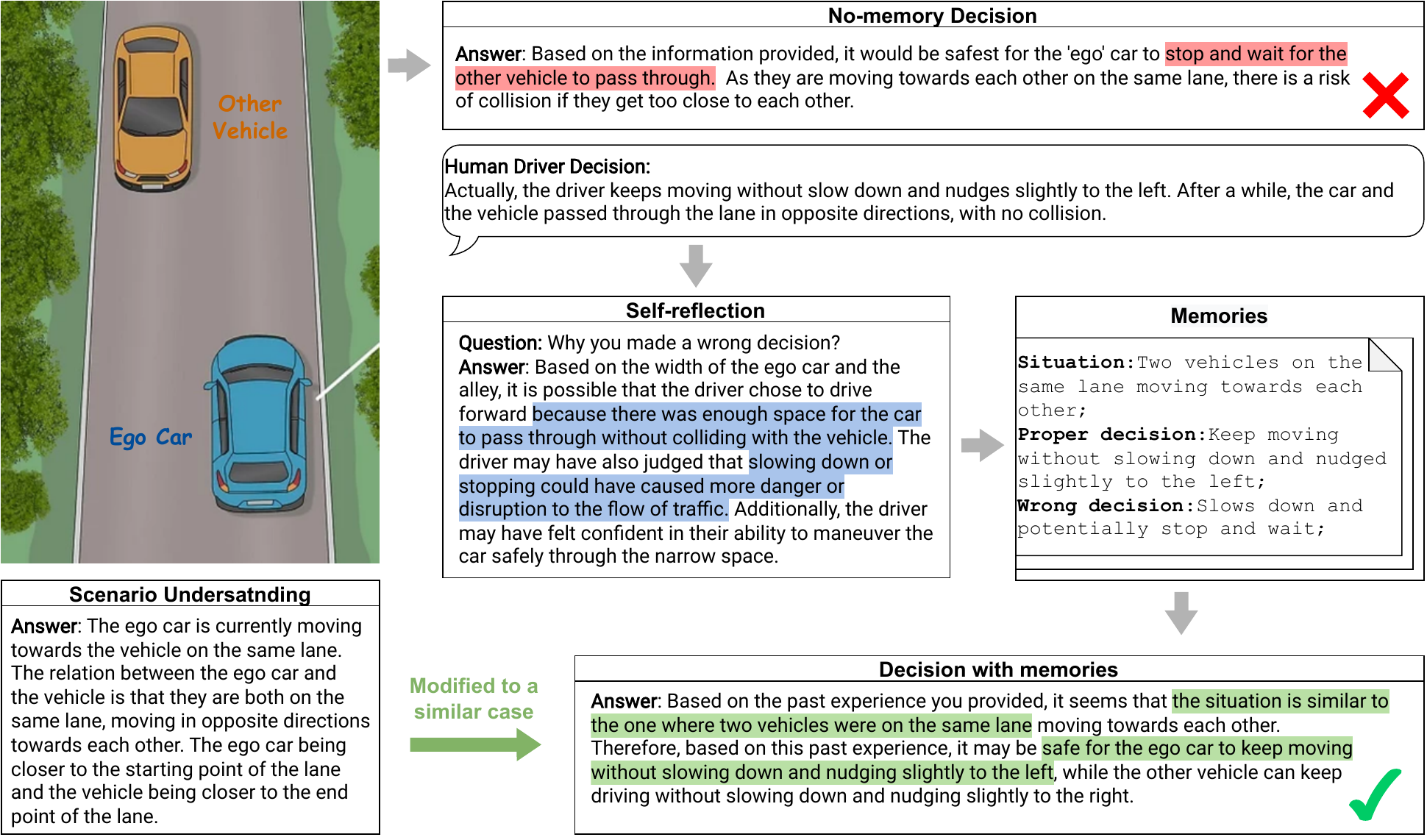}
    \caption{An example of the self-reflection and memorization ability of LLM}
    \label{fig:enter-label}
\end{figure}

As depicted in Fig.\ref{fig:motivation}(b), our proposed LLM-based approach closely follows the human through the use of a separate memory module. The memory module records only the decision scenarios that "deviate from the experts." The experts can be either the developer's evaluation of the LLM decision or the ground truth of human driver's decision in the real world.
Once expert feedback is obtained, the LLM undergoes a self-reflection process to determine why its decision deviates from the expert's. It then summarizes the traffic situation into a decision scenario and adds it to the memory pool as a new memory entry, along with the proper decision. When the next time a similar case is encountered, the LLM can quickly retrieve this memory entry for reference and make an informed decision.

Fig.\ref{fig:enter-label} gives an example of the memorization process. The scenario involves an encounter between a blue ego car and a yellow car travelling in opposite directions in a narrow lane that is slightly wider than twice the width of the car. After converting the scene into structured text for input into GPT-3.5, we find that the model understood the scene well, including the states, orientation and destination of the vehicles. However, when we asked it to make a decision about the scenario, GPT-3.5 gave a safe but overly cautious advice that the ego car should stop and wait for the other vehicle to pass first.
To improve the LLM's performance, the expert gives practical advice on how a human driver would handle the situation, which involves keeping the car moving and nudging it slightly to the left. The LLM then recognises that there is enough space for both vehicles to pass and that slowing down could disrupt the flow of traffic. It summarizes the situation as "two vehicles in the same lane moving towards each other" and records the memory along with the proper decision.
Using these memories, we input another scenario about two cars meeting in a narrow alleyway at different speeds and positions, and asked the LLM to make a decision. The LLM successfully recognises that this is just another variant of the "two cars in the same lane moving towards each other" decision scenario, and advises that it is safe for the ego car to keep going instead of slowing down and waiting.

The memorization ability continuously collects driving cases to gain experience and assists decision-making by retrieving existing memory, endowing LLM with the capability of continuous learning in the field of autonomous driving. Furthermore, this significantly reduces the decision cost of LLM in similar scenarios and improves its practical performance.

\newpage
\section{Related work}
\textbf{Self-driving Autonomy.}
Autonomous vehicles encompasses two main paradigms: modular~\cite{thrun2006stanley,daudelin2018integrated} and end-to-end~\cite{casas2021mp3,uniad,sadat2020perceive,tampuu2020survey}. The modular approach involves a stack of interconnected components that handle various sub-tasks, such as perception~\cite{li2023logonet,li2022bevformer,yin2021center}, planning~\cite{kelly2003reactive,zhang2022rethinking}, and control~\cite{peng2018sim,johnson2005pid}. This architecture offers enticing features like modularity and versatility. However, tuning the pipeline and managing error propagation can pose challenges. In contrast, end-to-end autonomy directly maps sensor input to planner or controller commands. These methods are typically easier to develop but lack interpretability, making it difficult to diagnose errors, ensure safety, and incorporate traffic rules. Nevertheless, recent advancements in end-to-end learnable pipeline autonomy have shown promising outcomes by combining the strengths of both paradigms~\cite{uniad,casas2021mp3}. Despite significant progress being made in these two paradigms of self-driving, there are often observed that they become fragile when handling long-tail data or out-of-distribution scenarios that occur in a real-world environment~\cite{kong2023robo3d}, which poses challenges to safety-critical autonomous driving.

\textbf{Advanced tasks with Large Language Models.}
The success of Large Language Models (LLM) is undoubtedly exciting as it demonstrates the extent to which machines can learn human knowledge. Recent efforts in LLM have shown impressive performance in zero-shot prompting and complex reasoning~\cite{bian2023chatgpt,nay2022law,chowdhery2022palm,ouyang2022training,chung2022scaling}, embodied agent research~\cite{wang2023voyager,zhu2023ghost,vemprala2023chatgpt,driess2023palm,yao2022react} and addressing key transportation problems~\cite{zheng2023chatgpt}. PaLM-E ~\cite{driess2023palm} employ fine-tuning techniques to adapt pre-trained LLM for supporting multimodal prompts. Reflexion~\cite{shinn2023reflexion} incorporates self-reflection to further enhance the agent's reasoning abilities with the chain-of-thought prompting~\cite{ouyang2022training} to generate both reasoning traces and task-specific actions using LLM. VOYAGER~\cite{wang2023voyager} presents lifelong learning with prompting mechanisms, skill library, and self-verification, which are based on LLM. These three modules aim to enhance the development of more complex behaviors of agents. Generative Agents~\cite{park2023generative} employ LLM to store complete records of an agent's experiences. Over time, these memories are synthesized into higher-level reflections and dynamically retrieved to plan behavior. 
% An easy idea of if the LLM also can be applied to the development of autonomous driving. The outstanding generalization and reasoning capability of LLM holds the potential to address and explain corner cases and rare events that occur infrequently in autonomous driving but are possible during human-driven scenarios. In this paper, we try to employ the LLM in autonomous driving and verify their ability to address the aforementioned problems.
Instruct2Act~\cite{huang2023instruct2act} introduces a framework that utilizes Large Language Models to map multi-modal instructions to sequential actions for robotic manipulation tasks.

\section{Conclusion}

In this paper, we present our idea of building a system that can drive like a human. We reckon that previous optimization-based autonomous driving systems have their limits when dealing with long-tail corner cases due to the catastrophic forgetting of global optimization. Therefore, we summarize three necessary abilities that an AD system should have to defeat imperfections including (1) Reasoning, (2) Interpretation, and (3) Memorization. Then we design a new paradigm following these three creeds that mimics the process of human learning to drive. Finally, with the hope of a primary artificial general intelligence, we try to use GPT-3.5 as our LLM test-bed and show an impressive ability of understanding traffic scenarios. 
As a preliminary work, we have only scratched the surface of LLMs' potential in closed-loop driving to highlight the benefits and opportunities of adopting this technology, rather than using the LLMs as a driving agent. Our aspiration is that this research will serve as a catalyst for both academia and industry to innovate and construct an AGI-based autonomous driving system that can drive like a human.
% \section{Submission of papers to NeurIPS 2023}

\medskip
{
\small
\bibliographystyle{plain}
\bibliography{main.bib}
}
\end{document}